\documentclass[10pt,twocolumn,letterpaper]{article}

\usepackage{wacv}
\usepackage{times}
\usepackage{epsfig}
\usepackage{graphicx}
\usepackage{amsmath}
\usepackage{amssymb}

\usepackage{array}
\newcolumntype{P}[1]{>{\centering\arraybackslash}p{#1}}

\newcommand{\vshrink}[0]{\vspace{-0.1cm}}

\newlength{\w}

\usepackage[pagebackref=true,breaklinks=true,letterpaper=true,colorlinks,bookmarks=false]{hyperref}

\wacvfinalcopy % *** Uncomment this line for the final submission

\def\wacvPaperID{63} % *** Enter the wacv Paper ID here

\ifwacvfinal\fi
\begin{document}

%%%%%%%%% TITLE
\title{Learning to Detect Multiple Photographic Defects}

\author{Ning Yu\hspace{0.01in}\textsuperscript{1}\hspace{-1ex}
\and Xiaohui Shen\hspace{0.01in}\textsuperscript{2}\hspace{-1ex}
\and Zhe Lin\hspace{0.01in}\textsuperscript{2}\hspace{-1ex}
\and Radom{\' i}r M{\v e}ch\hspace{0.01in}\textsuperscript{2}\hspace{-1ex}
\and Connelly Barnes\hspace{0.01in}\textsuperscript{1}\\
\hspace{-5in}		% If this is any larger then the first line wraps
\hspace{0.2in}\textsuperscript{1}\hspace{0.01in}University of Virginia \hspace{1.35in}
\hspace{0.15in} \textsuperscript{2}\hspace{0.01in}Adobe Research \hspace{-4.98in}\\
\hspace{-4.7in}		% If this is any larger then the first line wraps
{\tt\small \{ny4kt, connelly\}@cs.virginia.edu}
{\tt\small \hspace{0.5in} \{xshen, zlin, rmech\}@adobe.com}\\
}

\maketitle
\ifwacvfinal\thispagestyle{empty}\fi

%%%%%%%%% ABSTRACT
\begin{abstract}
In this paper, we introduce the problem of simultaneously detecting multiple photographic defects. We aim at detecting the existence, severity, and potential locations of common photographic defects related to color, noise, blur and composition. The automatic detection of such defects could be used to provide users with suggestions for how to improve photos without the need to laboriously try various correction methods. Defect detection could also help users select photos of higher quality while filtering out those with severe defects in photo curation and summarization.

To investigate this problem, we collected a large-scale dataset of user annotations on seven common photographic defects, which allows us to evaluate algorithms by measuring their consistency with human judgments. Our new dataset enables us to formulate the problem as a multi-task learning problem and train a multi-column deep convolutional neural network (CNN) to simultaneously predict the severity of all the defects. Unlike some existing single-defect estimation methods that rely on low-level statistics and may fail in many cases on natural photographs, our model is able to understand image contents and quality at a higher level. As a result, in our experiments, we show that our model has predictions with much higher consistency with human judgments than low-level methods as well as several baseline CNN models. Our model also performs better than an average human from our user study. 
\end{abstract}
\vspace{-0.5cm}
%%%%%%%%% BODY TEXT
\vshrink{}
\section{Introduction}

\begin{figure}[!t]
\centering
\includegraphics[width=1\linewidth]{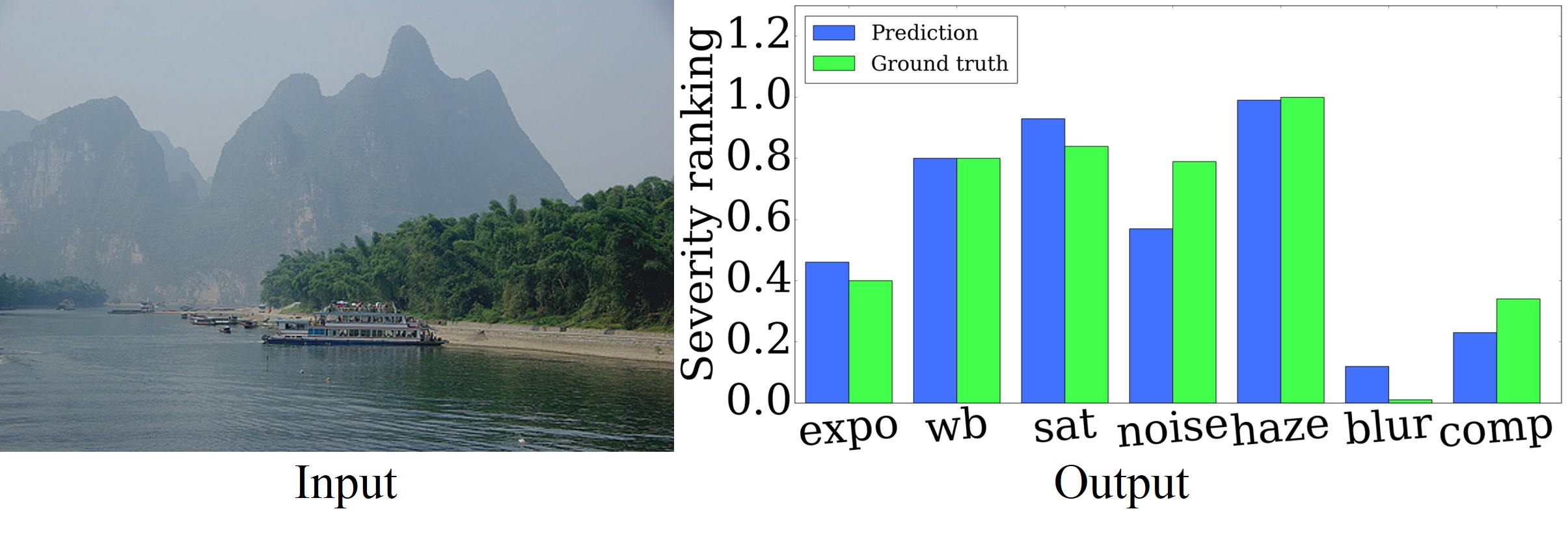}
\caption{An illustration of detecting multiple photographic defects. For each defect (from left to right: \textit{bad exposure}, \textit{bad white balance}, \textit{over/under saturation}, \textit{noise}, \textit{haze}, \textit{undesired blur}, \textit{bad composition}), we report the relative ranking of a severity score in percentage, compared to all the other photos in a testing set. Higher numbers indicate more severe defects. Our prediction rankings (blue) are consistent with the human judgment (green).}
\label{teaser}
\vspace{-12pt}
\end{figure} 

Many natural photos suffer from certain types of photographic defects, e.g., bad exposure, severe noise, and camera shake, due to imperfect capture conditions or limited expertise of the photographer. To improve those images, various manual tools in image editing software (e.g., Adobe Photoshop$^{\tiny{\textregistered}}$) and automatic adjustment methods in the research community have been developed to fix specific types of defects~\cite{ovsiannikov2010backlit,dabov2007image,he2011single,sun2013edge,fang2014automatic}. Because many factors affect image quality and there are abundant corrections available for each factor, it becomes difficult for a user without much photographic expertise to understand the defects in an image and choose proper correction methods. Moreover, with the explosion in the number of photos in one\rq s personal collection, it is also very tedious, if not impractical, for a user to go through all the photos and choose different corrections. It is therefore desirable to have a tool that can quickly identify the common defects in an image, suggest corresponding tools or auto-correction methods, and guide users to improve their photos. Furthermore, such a technology can also be applied in photo curation and collage to suggest good photos while filtering out bad ones.

To this end, we introduce the problem of simultaneously detecting multiple photographic defects. That is, we detect the existence and severity of a number of common photographic defects. By consulting professional photographers and analyzing a large amount of image editing data, we identified the seven most common defects, namely, \textit{bad exposure}, \textit{bad white balance}, \textit{over/under saturation}, \textit{noise}, \textit{haze}, \textit{undesired blur}, and \textit{bad composition}. Given a natural photo, we would like to predict the severity of these seven defects at the same time, as illustrated in Figure~\ref{teaser}. We note that although there is research on estimating the degree of specific defects (e.g., noise level~\cite{liu2013single} or blur amount~\cite{chakrabarti2010analyzing}), to our knowledge, there is no prior work addressing the problem of simultaneous detection of multiple defects.

\begin{figure}[!t]
\small
\centering
\setlength{\w}{1.3in}
\begin{tabular}{P{0.46\linewidth}|P{0.46\linewidth}}
(a) Noise & (b) Undesired blur \tabularnewline
\includegraphics[width=\w]{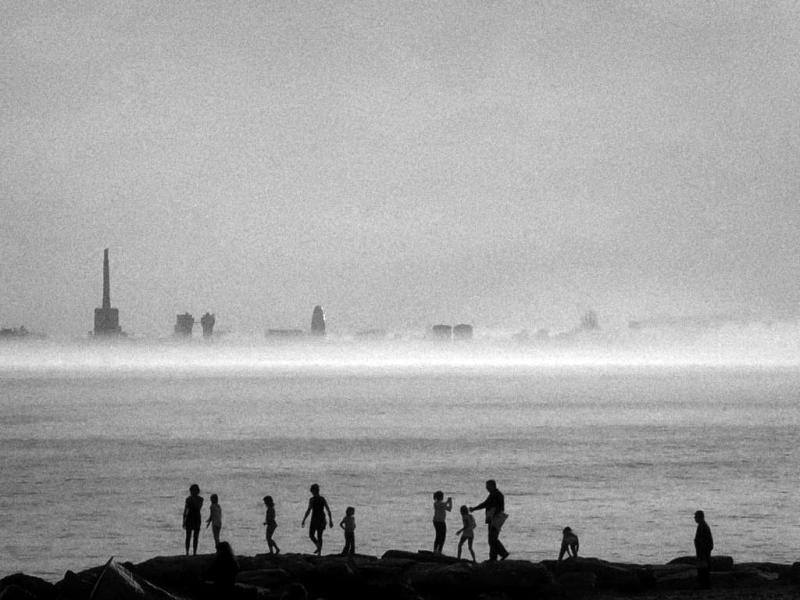} &  \includegraphics[width=\w]{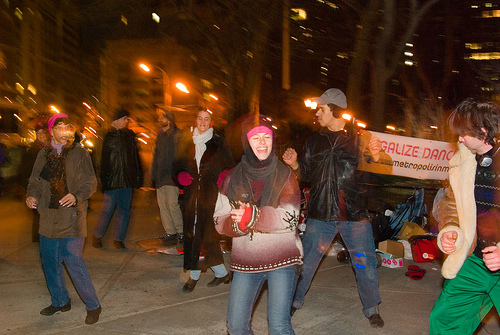} \tabularnewline
\cite{liu2013single}: $0\%$ &~\cite{chakrabarti2010analyzing}: $62\%$ \tabularnewline
Ours: $96\%$ & Ours: $98\%$ \tabularnewline
Ground truth: $98\%$ & Ground truth: $98\%$ \tabularnewline
\includegraphics[width=\w]{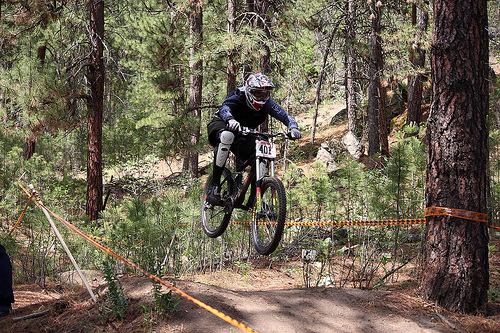} & \includegraphics[width=\w]{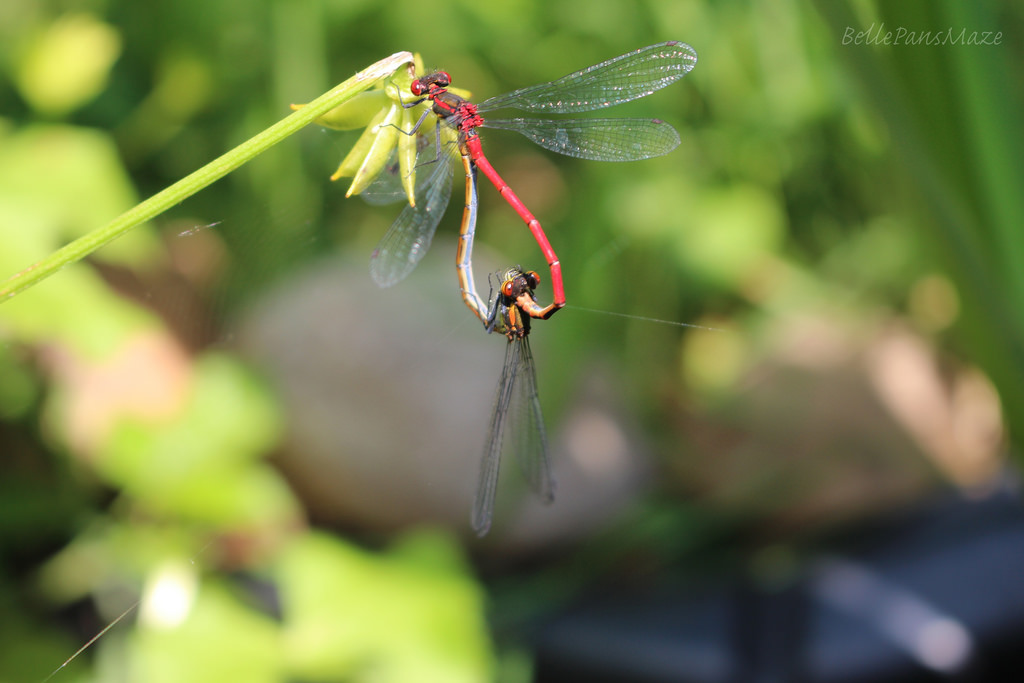} \tabularnewline
\cite{liu2013single}: $100\%$ &~\cite{chakrabarti2010analyzing}: $100\%$ \tabularnewline
Ours: $45\%$ & Ours: $4\%$ \tabularnewline
Ground truth: $60\%$ & Ground truth: $0\%$ \tabularnewline
\end{tabular}
\caption{Failure cases of the two previous defect estimation methods on \textit{noise}~\cite{liu2013single} and \textit{undesired blur}~\cite{chakrabarti2010analyzing}, respectively. Previous methods fail to detect the defects in the first row, and are confused by highly textured areas or desired depth-of-field in the second row. Our predictions are more consistent with ground truth. The percentage numbers measure the ranking of the image in terms of defect severity compared to other photos in a testing set.}
\label{baseline_failures}
\vspace{-12pt}
\end{figure}

To facilitate the research on this problem, we collected a dataset containing $12,853$ natural images, each with detailed user annotations on the severity of all the seven defects. This dataset allows us to train and evaluate algorithms based on human judgments, which is a distinct difference from previous methods that use synthesized artifacts as ground-truth defects~\cite{liu2013single, kang2014convolutional}. Synthetic defects are usually generated under certain assumptions regarding the defect patterns (e.g., Gaussian noise, or uniformly darkening the images). As a result, the methods developed on top of such data cannot cover the much more diverse defect patterns present in natural photos. For example in Figure~\ref{baseline_failures} (a), the noise level estimation method~\cite{liu2013single} is easily fooled by real noise or highly textured areas. Moreover, human judgments of image defects also heavily rely on understanding the important content that was intended to be captured. Without such higher-level understanding, the blur analysis method~\cite{chakrabarti2010analyzing} in Figure~\ref{baseline_failures} (b) cannot differentiate between undesired motion blur and a desired depth-of-field effect.

By contrast, leveraging the newly collected dataset, we formulate the problem as a multi-task prediction and learn a multi-column deep convolutional neural network (CNN) to simultaneously predict the severity of all the defects. By taking both the entire image and local patches as input, the learned model can better understand the image content while still being able to focus on local statistics, and have more accurate predictions, as shown in Figure~\ref{baseline_failures}.

The contributions of this paper are therefore: (1) We introduce a new problem of detecting multiple photographic defects, which is important for applications in image editing and photo organization. (2) We collect a new large-scale dataset with detailed human judgments on seven common defects, which will be released to facilitate the research on this problem. (3) We make a first attempt to approach this problem by training multi-column neural networks that consider both the global image and local statistics. We show in our experiments that our model achieves higher consistency with human judgments than previous single-defect estimation methods as well as baseline CNN models, and performs better than an average user.

Code and dataset are publicly available at \url{https://github.com/ningyu1991/DefectDetection}.

\vshrink{}
\section{Related Work}
We discuss previous work, as grouped into three areas.
\vspace{-9pt}
\paragraph{Single defect estimation and correction.}  There have been many efforts focused on fixing a specific type of photographic defect, e.g., exposure correction~\cite{russ1995image,yuan2012automatic}, haze removal~\cite{he2011single,zhu2015fast,ren2016single}, denoising~\cite{paris2006fast,dabov2007image,adams2009gaussian}, deblurring~\cite{ fergus2006removing,sun2015learning} and image cropping~\cite{liu2010optimizing,yan2013learning,fang2014automatic}. However, most of these methods directly generate an improved image without explicitly estimating the existence or severity of the defect. The level of white Gaussian noise in an image is explicitly estimated in~\cite{liu2013single}, while Chakrabarti \textit{et al.}~\cite{chakrabarti2010analyzing} analyze the amount of spatially-varying blur. Both of these methods rely on low-level statistics under the assumption that such defects already exist in the image, and may not work very well given an arbitrary natural photo. More importantly, none of those previous works tackles the detection of all the common defects at the same time, as in our study.
\vspace{-9pt}
\paragraph{Deep convolutional neural networks (CNN).} Deep convolutional neural networks~\cite{szegedy2015going,he2015deep} have shown tremendous success in capturing high-level image content, and have achieved state-of-the-art in various computer vision tasks~\cite{ioffe2015batch,chatfield2014return,dixit2015scene}. Previous papers have demonstrated that multi-column networks can have improved performance over single-column networks~\cite{ciregan2012multi,lu2014rapid,lu2015deep}, by leveraging the information from multiple related tasks, or taking inputs with different scales~\cite{oquab2014weakly,chen2015attention,lu2015deep}. Inspired by these results, we formulate the multiple defects estimation problem as a multi-task prediction, and design an end-to-end multi-column network that shares weights in earlier stages and splits out columns in the later stages for each defect.
\vspace{-9pt}
\paragraph{Image quality assessment.} The conventional no-reference image quality assessment (NR-IQA) evaluates visual distortions including JPEG compression, additive white Gaussian noise, Gaussian blur, etc.~\cite{li2011blind,chetouani2010novel,kang2014convolutional}. In these tasks, distortions are synthetically added and uniformly distributed over the entire image. On the contrary, our problem focuses on common defects found in photos in the wild, which exist mainly due to limitations at capture time. Our problem does not have any assumptions regarding the existence, types, and locations of the defects, and involves high-level image content understanding driven by human judgment, and is therefore a significantly different problem. The recent work of deep photo aesthetics assessment~\cite{mai2016composition} directly classifies query images into high or low aesthetics, which is also different from our problem. %However, our defect detection results may potentially add latent reasonings for their aesthetics assessment with a finer granularity from binary classification to continuous prediction.

\vshrink{}
\section{Photographic Defect Severity Dataset}

Because our problem is new and involves human judgments, we need to run a user study to collect human judgments, and also define a suitable evaluation metric. We first discuss in Section~\ref{Data Collection} our new dataset with detailed human annotations on natural photos. Next, in Section~\ref{Evaluation Metric}, we introduce an evaluation metric that is well-suited for our problem. Then, in Section~\ref{User Consistency Analysis}, we provide user consistency analysis based on the proposed evaluation protocol. We will release our dataset and evaluation protocol to promote research on this problem.

\vshrink{}
\subsection{Data Collection}
\label{Data Collection}

To determine the most common and important defects, we consulted professional photographers and analyzed a large amount of image editing data. In the end, we selected seven types of photographic defects: \textit{bad exposure}, \textit{bad white balance}, \textit{over/under saturation}, \textit{noise}, \textit{haze}, \textit{undesired blur}, and \textit{bad composition}.

We then randomly sampled $12,853$ natural photos from the Yahoo Flickr Creative Commons 100M dataset~\cite{thomee2015new}, and obtained the annotations of severity scores for each defect through Amazon Mechanical Turk (AMT)\footnote{\url{www.mturk.com}}. Specifically, for the \textit{over/under saturation} defect, we provided five levels of severity for users to choose from: $\{$\textit{severely under-saturated}, \textit{mildly under-saturated}, \textit{normal saturation}, \textit{mildly over-saturated}, \textit{severely over-saturated}$\}$, which map to a score set of $\{-1.0, -0.5, 0.0, 0.5, 1.0\}$. For all the other defects, we provided three levels of severity: $\{$\textit{none}, \textit{mild}, \textit{severe}$\}$, which map to a score set of $\{0.0, 0.5, 1.0\}$. 

When collecting the annotations, we randomly inserted a small set of ``sanity check" images with known defect severity levels. Most of these images have obviously severe defects or are defect-free, so a careful user will do a very good job on these images. We can thus filter out bad user annotations by measuring users' performance on those images. More details about the data collection process, such as the user interface, the qualification test, and the quality control procedure, are included in the supplementary material.

In the end, each image has five valid user annotations for each defect. We calculate the final ground-truth severity scores as a  weighted average over the five user annotations, in which the weights are proportional to users' accuracy on the ``sanity check" images and normalized among the five users. We found that such a weighted averaging process can significantly reduce annotation noise, and generate quite consistent ground-truth scores. More analysis regarding users' consistency is described in Section~\ref{User Consistency Analysis}. 

\begin{figure}[!t]
\centering
\includegraphics[width=1\linewidth]{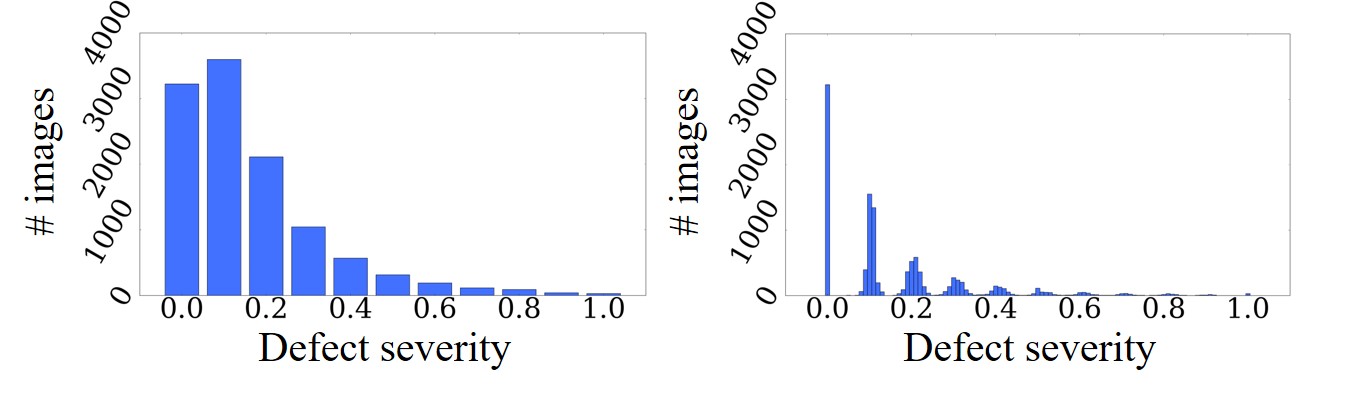}
\caption{The histogram of the ground-truth severity scores for the \textit{bad exposure} defect, shown with a coarse bin size (left) and with a fine bin size (right). Each bin in the left figure corresponds to a peak in the right figure. The hiogram has 11 peaks due to the discrete levels used in user annotation.}
\label{two_histograms}
\vspace{-12pt}
\end{figure}

Figure~\ref{two_histograms} (left) shows the distribution with a coarse bin size of the ground-truth severity scores for the \textit{bad exposure} defect. The distribution has a long tail, with most images containing no or mild exposure problems. This is expected since the images are randomly sampled from a large collection of photos and follow a natural distribution in terms of image quality. Another interesting observation is that the scores, even if plotted with a fine bin size (as shown at right in Figure~\ref{two_histograms}), form 11 peaks. That is because the annotations given by each user have three levels: $0.0$ (\textit{none}), $0.5$ (\textit{mild}), and $1.0$ (\textit{severe}). An equal weighted average over five user annotations would result in 11 discrete levels from $0.0$ to $1.0$ with a step of $0.1$. When the averaging is weighted by users' accuracy, the scores become a little more dispersed but still form 11 peaks around those discrete levels. Each peak in the right histogram corresponds to a coarse bin in the left histogram. We observe a similar distribution for the scores of other defects except for saturation, whose score distribution has 21 peaks, because its annotations by each user have five different levels instead of three.

Finally for experimental evaluation, we randomly split the dataset into a training set with $11,313$ images and a testing set with $1,540$ images.

\begin{table*}[!t]
\small
\centering
\caption{The mean cross-class $\rho$ and Kendall's $W$ among user annotations for each defect.}
\begin{tabular}{P{3.1cm}||P{1.2cm}|P{1.4cm}|P{1.3cm}|P{0.9cm}|P{0.9cm}|P{1.3cm}|P{1.55cm}||P{0.9cm}}
\hline
& Bad & Bad white & Over/under & Noise & Haze & Undesired & Bad & Mean\tabularnewline
& exposure & balance & saturation & & & blur & composition & \tabularnewline
\hline\hline
Cross-class $\rho$ & 0.7691 & 0.7498 & 0.7944 & 0.8236 & 0.8530 & 0.8528 & 0.6168 & 0.7799\tabularnewline
Kendall's $W$ & 0.5247 & 0.4863 & 0.5435 & 0.5470 & 0.6208 & 0.6388 & 0.4203 & 0.5402\tabularnewline
\hline
\end{tabular}
\label{consistency}
\end{table*}

\vshrink{}
\subsection{Evaluation Metric}
\label{Evaluation Metric}

In order to fairly evaluate the performance of users and algorithms, it is important to have an evaluation metric that is suitable for our problem and dataset. We first discuss three key properties that a metric should have, then discuss limitations of some simple evaluation metrics, and next discuss our preferred metric.

The three key properties that we desire from an evaluation metric are: (1) \emph{Balance:} the metric should give a roughly equal contribution to the final cost for images that fall under each severity of defect. This is because there are much more defect-free images than defective ones in our dataset, as shown in Figure~\ref{two_histograms}, so the evaluation metric should perform a rebalancing to account for this; (2) \emph{Proportionality:} the metric should consider slight errors in prediction as better than extreme errors in prediction. For example, if we have a defect-free image (class 1) and we predict that it is slightly defective (class 2), this should be better than predicting that the same image is highly defective (class 11); and (3) \emph{Ranking:} a ranking-based metric that considers only the order of the predictions from the model is preferable to an absolute metric. This is especially true for applications where the relative ranking is important, such as photo ranking or curation~\cite{kong2016photo}, and for comparisons with previous work, where the scores output by an algorithm may not be directly comparable to the user severity scale.

We now discuss how a few simple metrics fall short of the key properties. The $L^2$ loss does not satisfy key properties of balance (1) and ranking (3). The overall classification accuracy could be computed by quantizing the defect scores into 11 or 21 classes as discussed in Section~\ref{Data Collection}. However, accuracy does not satisfy any of the key properties. The classification accuracy given varying class bias tolerances is generalized to satisfy proportionality (2) but still does not satisfy the other properties. A Spearman Rank Correlation Coefficient~\cite{myers2010research} could be computed by forming two ranked image lists based on the prediction and ground-truth scores, respectively. However, the Spearman Rank Correlation does not satisfy the key property of balance (1) and proportionality (2). In particular, it fails at proportionality (2), because two sets of images that all fall into a given class such as slightly defective (class 2) can still have quite different rankings. % Connelly: removed due to error after discussion with Ning

In order to satisfy all three key properties, in this work we propose a new evaluation metric, the \textbf{cross-class ranking correlation} (cross-class $\rho$). Specifically, we assign the test images to one of the 11 classes according to their ground-truth defect severity scores (21 classes for saturation). The classes (the bins in Figure~\ref{two_histograms} left) naturally fit the peaked distribution of our dataset, which is shown in Figure~\ref{two_histograms} right. During evaluation, we randomly sample one image from each class from the ground truth. Those images compose an ordered list based on the severity levels of classes they are sampled from. When a prediction is made for the defect scores of those images, we can also sort the images according to their predicted scores and form another ordered list. We then calculate the regular Spearman Rank Correlation Coefficient $\rho$~\cite{myers2010research} between the two lists, yielding a score within $[-1,1]$. A larger Spearman coefficient indicates the orders in the two lists are more similar, and the predictions are more consistent with the ground truth. To obtain a robust evaluation, we repeat the random sampling and correlation calculation many times ($15,000$) and use the mean as our final cross-class $\rho$.% \note{Need to edit this paragraph to connect our metric to the peaked distribution of our data, and also make the definition not just be for algorithms but also humans.}

With the cross-class ranking correlation, we achieve all three key properties of an evaluation metric. During image sampling, each class only contributes one image to the list, so this acts to rebalance the dataset, and satisfy the balance property (1). The property of proportionality (2) is satisfied because no penalty is applied if two images fall within the same class, and if images are within a different class, the correlation decreases as the classes become further apart. The ranking property (3) is trivially satisfied.

\vshrink{}
\subsection{User Consistency Analysis}

\label{User Consistency Analysis}

After specifying our evaluation metric, we are able to examine the consistency of AMT users' annotations. We conduct consistency analysis on each group of five users who annotated the same batch of images. We compare the annotations from two users against the other three annotations. Specifically, we calculate the mean annotations among the two subgroups separately and utilize the cross-class $\rho$ to evaluate the consistency. For each batch, we average the correlations over all possible two-against-three splits. We additionally estimate the p-value of a t-test for each correlation, which measures the statistical significance of the correlation relative to a null hypothesis of uncorrelated response. We use the Benjamini-Hochberg procedure~\cite{benjamini2001control} to control the false discovery rate (FDR) for multiple correlation hypotheses. At an FDR level of $0.05$, we calculate the percentage of batches with significant agreement among users. The average cross-class $\rho$ for each defect are listed in Table~\ref{consistency}. They are all above $0.6$, and mostly around $0.8$, where the valid range for $\rho$ is $[-1,1]$. The percentage of significant batches is at least $99.55\%$ for all the defects.% This indicates that the user annotations are quite consistent.%, and the measurements are statistically significant. 

We further evaluate the annotation consistency with Kendall's Coefficient of Concordance ($W$)~\cite{kendall1939problem}, which directly calculates the agreement among multiple users, and accounts for tied ranks. Kendall's W ranges from $0.0$ (no agreement) to $1.0$ (complete agreement). We estimate the p-value of a Chi-squared test to evaluate the statistical significance. We use the same Benjamini-Hochberg procedure to measure the percentage of batches with the significant agreement. Kendall's $W$ values for each defect are listed in Table~\ref{consistency}. These show a similar trend as cross-class $\rho$, and also have a percentage of batches with significant agreement of at least $99.55\%$. Both measures demonstrate the consistency across AMT users and indicate that the annotations are reliable for scientific research.

\vshrink{}
\section{Simultaneous Detection of Multiple Defects}

The availability of this new dataset enables us to train a deep convolutional neural network (CNN) to directly learn high-level understanding of photographic defects from human judgments. In this section, we describe the details of CNN training, including the  architecture, pre-processing of input images, loss functions, and the data augmentation process to rebalance our  skewed training data.

\begin{figure}[!t]
\centering
\includegraphics[width=1\linewidth]{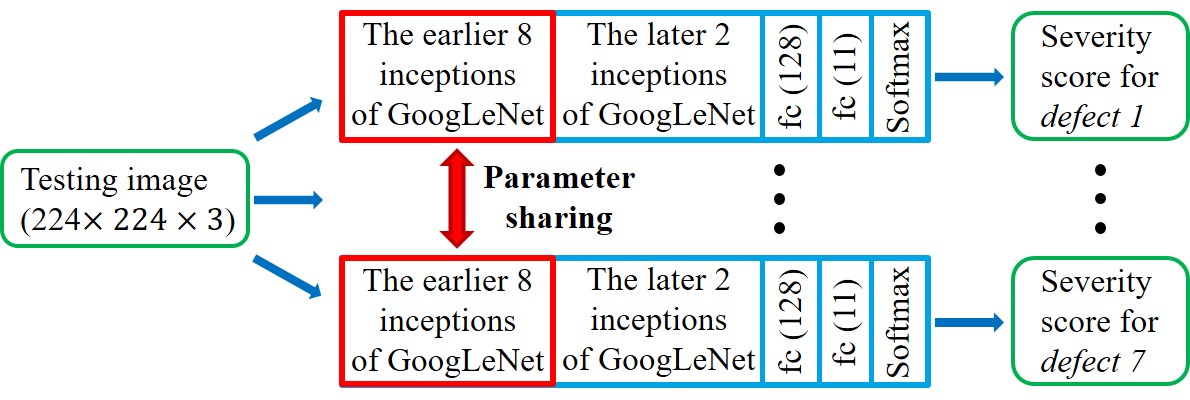}
\caption{This diagram shows the multi-column GoogLeNet~\cite{ioffe2015batch} architecture for multi-task predictions. Here ``fc ($x$)" represents a fully connected layer with $x$ hidden neurons. Red frames indicate the layers with shared parameters.}
\label{holistic_patch_input_model}
\vspace{-12pt}
\end{figure}

\vshrink{}
\subsection{Multi-Column Network Architecture}

Our goal is to predict the severity of seven defects at the same time. These defects are related to low-level photo properties such as color, exposure, noise, and blur, and high-level properties such as faces, humans, and compositional balance. We note that both low- and high-level content features may be useful. Therefore, we use a multi-column CNN, in which the earlier layers of the network are shared across all the defects to learn defect-agnostic features, and in later layers, a separate branch is dedicated to each defect to capture defect-specific information. Figure~\ref{holistic_patch_input_model} shows our architecture. We build upon GoogLeNet~\cite{szegedy2015going,ioffe2015batch}, which contains convolutional modules called inceptions. We select GoogLeNet rather than other prevalent architectures, e.g., VGG nets~\cite{simonyan2014very} or ResNet~\cite{he2016deep}, because its lighter memory requirement enables multi-column training with a larger batch size. We use the first 8 inceptions of GoogLeNet~\cite{ioffe2015batch} as shared layers, and then dedicate a separate branch for each defect with two inceptions and fully-connected layers.

We also tried two other baseline models: (1) A single-column network that directly predicts seven defect scores, and (2) Seven separate networks, each predicting one defect. The comparisons in Section~\ref{ablation study} show that our branching architecture is better than these alternatives.%the shared information during joint training and specific modules for each defect in our multi-column network are both important. 

\vshrink{}
\subsection{Network Input}

In order to detect the defects, we need both a global view of the entire image and a focus on the statistics in local image areas. Therefore, we prepare two different types of inputs for the network: (1) downsized holistic images, which contains complete image content, and (2) patches randomly sampled from the images at the original resolution, which retains high-frequency statistics especially useful for the detection of some defects such as noise and blur. We make the simplifying assumption that we can assign each local patch the same holistic severity score. This sometimes introduces noise when a defect appears only locally. But by aggregating over many patches the network can learn to ignore this noise (this can be observed in Table~\ref{comparison_cross_class_correlation} in the outperformance of the patch model for certain defects). We  also tried a weakly supervised architecture by estimating an attention map (similar to~\cite{xu2015show}) which uses different weights for each patch, but the results show it did not help in our case.

Mixing the two input types would confuse the network during training. Accordingly, we train a separate network in Figure~\ref{holistic_patch_input_model} for each type of input, with a goal that each network can capture complementary information. 

The network with patch inputs does not predict the score regarding bad composition, because image composition should be solely considered over the entire image. For all the other six defects, we combine the predicted scores from the patch and holistic networks. We found that a simple average of scores using equal weights achieves good results, and outperforms both of the two individual models, as shown in Section~\ref{ablation study}. This demonstrates that the global and local information captured in the two networks is  complementary for this problem. We also tried to optimize the weights using quadratic programming on a separate validation set, but did not observe much improvement from this.

\vshrink{}
\subsection{Loss Functions}
\label{Loss Functions}

Due to the distribution of our ground truth annotations as shown in Figure~\ref{two_histograms}, where the scores are mostly distributed around discrete  peaks, we found that it works better to formulate our loss to involve classification rather than regression. However, the standard cross-entropy loss used in classification ignores the relations between the classes, as discussed in Section~\ref{Evaluation Metric}. In other words, all misclassifications are treated equally. In our case, we should impose more penalty if we misclassify an example in class 1 (no defect) to class 11 (severe defect), compared with a misclassification from class 1 (no defect) to class 2 (very mild defect). This is the property of proportionality from Section~\ref{Evaluation Metric}.

To accommodate such requirement, we use the infogain multinomial logistic loss\footnote{\url{http://caffe.berkeleyvision.org/doxygen/classcaffe_1_1InfogainLossLayer.html}} to measure the classification errors. The infogain loss $E$ is mathematically formulated as
\begin{equation}
E = -\frac{1}{N}\sum_{n=1}^N\sum_{k=1}^KH_{l_n,k}\log(p_{n,k}),
\label{infogain}
\end{equation}
where $N$ is the number of image samples; $K$ is the number of classes; $l_n$ is the class ground truth of the $n^{th}$ sample; $p_{n,k}$ is the probability of the $n^{th}$ sample classified to the $k^{th}$ class, which is the output after the softmax layer satisfying $\sum_k^Kp_{n,k}=1$ and $p_{n,k}\geq 0$. Finally, $H_{l_n,k}$ is the infogain weight for the $n^{th}$ sample with ground truth $l_n$ to be classified to class $k$. The higher the weight, the greater the reward for that classification result. Therefore, we can assign higher weights between similar classes. We derive our defect-specific infogain matrices from a naive conditional independence assumption and statistics of individual AMT users' case-by-case annotations. The details are included in the supplementary materials. During testing, once we obtain the classification probabilities over all the classes for an image, since each class is associated with a severity score, we can use the probabilities as weights to obtain an averaged severity score. That score is treated as our final prediction regarding the defect severity of the image.

Note that we use infogain loss only for training and prefer the cross-class $\rho$ for evaluation. This is because cross-class $\rho$ is a metric that uses ranking, and for applications such as photo curation, we care more about relative rankings than absolute scores.

We experimentally show in Section~\ref{ablation study} that training using the infogain class achieves better performance than using the standard cross-entropy loss. We also tried formulating the prediction as a regression task, and use $L^2$ loss compared with ground truth scores. The results are reasonable, but not as good as using the infogain loss. %We put a note here to explain why we do not directly use the infogain loss as our testing evaluation metric. The infogain loss considers the distance between two distributions, and is useful for training because it is differentiable when applied to a classification problem. However, for applications such as photo curation, given a testing image, we care more about its relative ranking, which is a deterministic scalar based on the score output by the network as opposed to being based on the probabilities output by the network. As an analogy, in e.g. a simple image classification task, it is common to use the cross-entropy loss for training (which is based on probabilities) but accuracy for testing evaluation (which is based on scores).
% Long version (removed to save length):
%We put a note here to explain why we do not directly use the infogain loss as our testing evaluation metric. The infogain loss considers the distance between two distributions, and is useful for training because it is differentiable when applied to a classification problem. However, for applications such as photo curation, given a testing image, we care more about its relative ranking, which is a deterministic scalar based on the score output by the network as opposed to being based on the probabilities output by the network. As an analogy, in e.g. a simple image classification task, it is common to use the cross-entropy loss for training (which is based on probabilities) but accuracy for testing evaluation (which is based on scores).

\vshrink{}
\subsection{Data Augmentation}
\label{Data Augmentation}
As discussed in Section~\ref{Data Collection}, our training data is heavily unbalanced with a high percent of defect-free images. In order to prevent the training from being dominated by defect-free images, we augment more training data on images with severe defects. This also  better satisfies the property of balance from Section~\ref{Evaluation Metric}. We augment samples in inverse proportion to class member counts but clamp the minimum and maximum sample counts to 5 and 50, respectively. The augmentation operation for the holistic input is random cropping (at half the receptive field) and warping, and for the patch input, random cropping. More details and the histograms before and after augmentation are shown in the supplementary material.  We experimentally validate in Section~\ref{ablation study} that our data rebalancing is crucial to the results.

\vshrink{}
\subsection{Implementation Details}
\label{Implementation Details}

The network is initialized from the GoogLeNet model~\cite{szegedy2015going} trained for ImageNet classification~\cite{deng2009imagenet}. We made some slight modifications on the architecture to make the model more compact and efficient: (1) we remove the two auxiliary classifier branches \textit{loss1} and \textit{loss2}, (2) we trim off the $3 \times 3$ convolution branch in \textit{inception\_5b}; and (3) in \textit{inception\_5b}, we reduce the number of output features of the $1 \times 1$, $3 \times 3$ double, and the pooling projection layers to $88$, $56$, and $32$, respectively. The output feature dimension of \textit{inception\_5b} is thus reduced from $1,024$ in the original network to $176$. 

During training, the batch size is $32$. The initial learning rate is $0.0001$ for the parameter-shared layers and is $10$ times larger for the defect-specific layers. All learning rates are multiplied by $0.96$ after every $6,400$ iterations. We set weight decay as $0.0002$ and momentum as $0.9$. We implement the training and testing in Caffe~\cite{jia2014caffe}.

During testing, to obtain the patch model predictions, we crop $K$ random patches from each image and average the scores from the patch networks. We set $K=10$, which gives a good trade-off between testing time and robustness.

%We share parameters of the earlier eight inceptions of GoogLeNet (from \textit{conv1/7$\times$7} to \textit{inception\_4e}) across all columns. We specify the last two inceptions (from \textit{inception\_5a} to \textit{pool/7$\times$}) for each column independently, and add one more fully connected layer (with $128$ hidden neurons) followed by a ReLU layer after \textit{pool/7$\times$7} to increase defect-wise capacity. We reduce the number of output neurons of the last fully connected layer from $1,000$ to $11$ (or $21$ for \textit{bad saturation}), and replace the multinomial logistic loss layer with a defect-specific infogain multinomial logistic loss layer (see Section~\ref{Loss Functions}). Infogain losses for all the defects have the same weight. The columns have identical initialization from the well-trained GoogLeNet model.

\vshrink{}
\section{Experiments}

\begin{figure}[!t]
\centering
\includegraphics[width=1\linewidth]{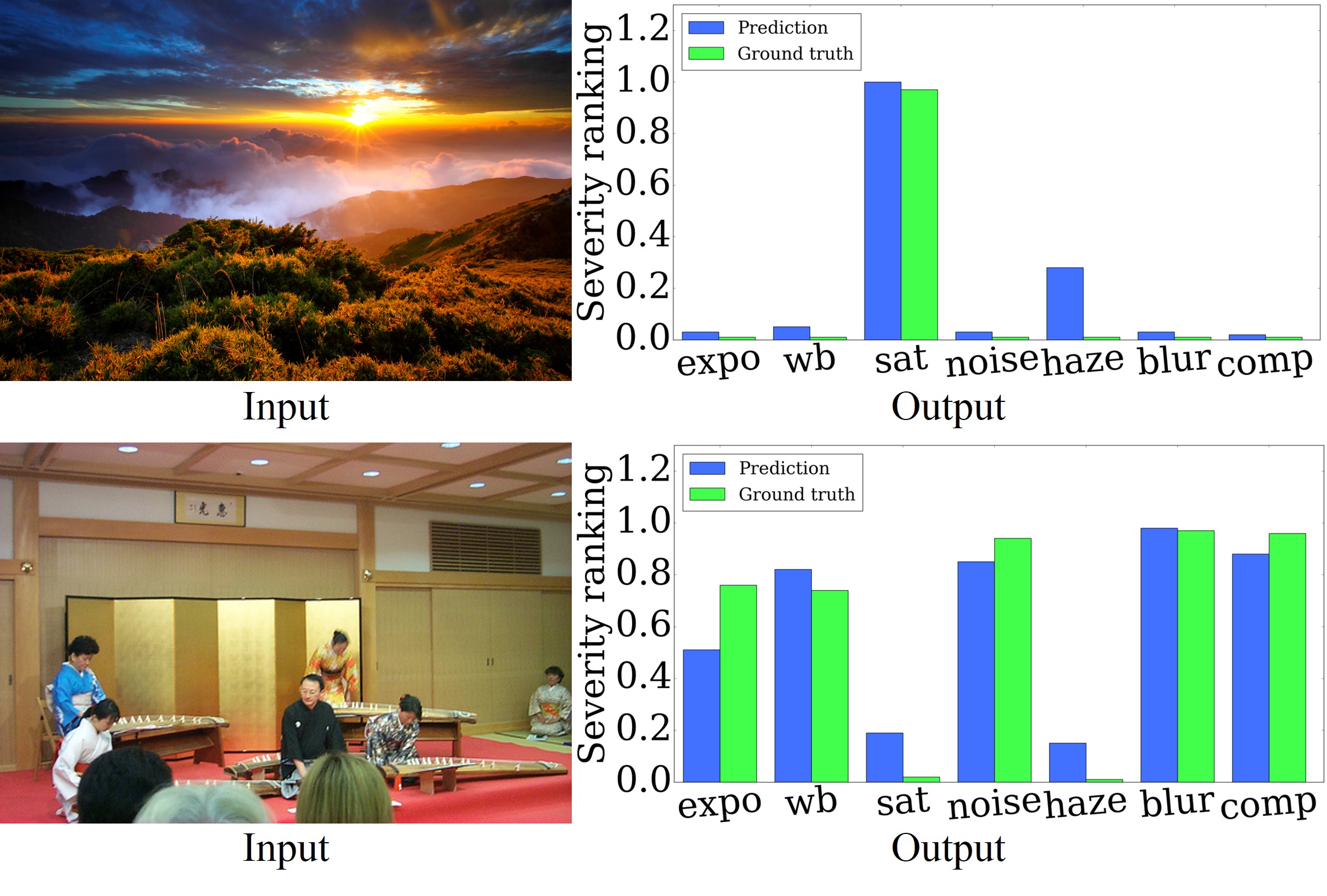}
\caption{Two visual results of our defect detection. For each defect (from left to right: \textit{bad exposure}, \textit{bad white balance}, \textit{over/under saturation}, \textit{noise}, \textit{haze}, \textit{undesired blur}, \textit{bad composition}), we report the relative ranking of a severity score in percentage, which measures the defect severity of a given image compared to all the other photos in a testing set. Higher numbers indicate more severe defects. Our prediction rankings (blue) are consistent with the human judgment (green).}
\label{results}
\vspace{-12pt}
\end{figure}

To predict all the seven defects, the  testing time of the proposed model on our Intel i7-6950X CPU (3.00GHz) is 3.6 sec. The average testing time on our NVIDIA Titan X GPU is about 0.5 sec. The holistic model requires 108 MB of memory and the patch model requires 97 MB. Two additional qualitative results are presented in Figure~\ref{results}. % We think both the computation burden and the model size are reasonable and can be used on mobile. But in future work, both speed and size could be further improved via deep model compression techniques (e.g.,~\cite{iandola2016squeezenet}).

\begin{table*}[!t]
\small
\centering
\caption{Comparison with baseline CNNs in terms of the cross-class $\rho$ on our testing dataset. \textbf{Bold} indicates the best performance. \underline{Underline} indicates the second best.}
\begin{tabular}{P{3.5cm}||P{1.2cm}|P{1.4cm}|P{1.3cm}|P{0.9cm}|P{0.9cm}|P{1.3cm}|P{1.55cm}||P{0.9cm}}
\hline
& Bad & Bad white & Over/under & Noise & Haze & Undesired & Bad & Mean\tabularnewline
& exposure & balance & saturation & & & blur & composition & \tabularnewline
\hline\hline
Multi-column (holistic) & 0.7529 & 0.7614 & 0.8996 & 0.6736 & 0.8346 & 0.6032  & \textbf{0.7123} & 0.7482\tabularnewline
Multi-column (patch) & 0.7825 & 0.8000 & 0.8923 & \textbf{0.8197} & 0.7759 & 0.6696 & - & -\tabularnewline
%QP combined & 0.8015 & \underline{0.8265} & \textbf{0.9113} & 0.7893 & \underline{0.8411} & 0.6923 & 0.7108 & \underline{0.7967} \tabularnewline
Multi-column (combined) & 0.8008 & \underline{0.8249} & \textbf{0.9098} & \underline{0.8174} & \textbf{0.8490} & 0.6867 & \textbf{0.7123} & \textbf{0.8001}\tabularnewline
\hline
Single-column architecture & \underline{0.8063} & 0.8201 & 0.8817 & 0.7246 & 0.7778 & \textbf{0.7447} & 0.5969 & 0.7646\tabularnewline
Separate networks & 0.7972 & 0.7925 & \underline{0.9039} & 0.7403 & 0.8315 & \underline{0.7209} & 0.6656 & 0.7788\tabularnewline
\hline
Regression $L^2$ loss & \textbf{0.8145} & \textbf{0.8323} & 0.8995 & 0.8118 & \underline{0.8394} & 0.7008 & 0.6169 & \underline{0.7879}\tabularnewline
Classification loss & 0.7850 & 0.7938 & 0.8969 & 0.7426 & 0.7867 & 0.7205 & \underline{0.6929} & 0.7740\tabularnewline
\hline
Without augmentation & 0.7864 & 0.7675 & 0.8907 & 0.7096 & 0.8076 & 0.6349 & 0.5383 & 0.7336\tabularnewline
\hline
\end{tabular}
\label{comparison_cross_class_correlation}
\end{table*}

\vshrink{}
\subsection{Ablation Study}
\label{ablation study}

\paragraph{Network Input.} The first three rows in Table~\ref{comparison_cross_class_correlation} show the cross-class $\rho$ of the multi-column network with holistic image input,  the network with patch input, and the combination of the two, respectively. The mean cross-class $\rho$ in the last column is obtained by averaging the values over all the seven defects. We can see that after combination, the results improve on almost all the defects. This shows the complementarity between the holistic and patch model. 

In all subsequent ablation studies, we use the same combined inputs for all the CNN models. That is, we separately train two networks for holistic images and local patches, and average the predictions with equal weight.
\vspace{-9pt}
\paragraph{Network architectures.} To investigate the necessity of having separate branches for each defect, we train a single-column network, in which the parameters are all shared for the defects except the last output.  To have a fair comparison in terms of model capacity, we increase the numbers of feature channels in the last two inceptions in the single-column network, to make the number of parameters for this model similar to our model. The results of the single-column network are reported in the 4th row in Table~\ref{comparison_cross_class_correlation}. We can see that the results on most defects become worse, as does the mean cross-class $\rho$. The performance on the composition defect has the biggest decrease, probably because understanding image composition requires higher-level features than color or texture, which are more important for other defects. 

On the other hand, one can train a separate network for each single defect, without sharing any parameters. To investigate this, we train a GoogLeNet for each defect separately and report the results in the 5th row in Table~\ref{comparison_cross_class_correlation}. We note that by unsharing the parameters, the number of overall trainable parameters in this model is much higher than the one in our multi-column network, resulting in much larger model size and longer testing time. However, our model has better performance on all defects except blur. We investigated the gaps between training and testing performance, and found that the separate networks for single defects are more prone to over-fitting, whereas the shared layers in our network act as a regularizer to improve generalizability.%are regularized by multiple defects and have better generalizability. 

After comparing with these  baseline CNN models, we find that our multi-column architecture gives a good trade-off between performance, compactness, and efficiency.
\vspace{-18pt}

\paragraph{Loss functions.} The results of the networks trained using $L^2$ loss as regression, and using cross-entropy loss as classification, are shown in the 6th and 7th rows in Table~\ref{comparison_cross_class_correlation}, respectively. This shows that using these two losses can also result in reasonably good performance. However, the infogain loss outperforms these two losses on the predictions for several defects as well as the overall mean cross-class $\rho$.

We realize we can reach the second best performance when we train using $L^2$ loss. Therefore, in order to the show the significance of the outperformance of our best model, we calculate the p value of the two-tailed Student’s t-test between the two networks. Here p is 0 to within the double precision accuracy, which indicates that training with the infogain loss  significantly outperforms training with $L^2$ loss.
\vspace{-27pt}

\paragraph{Data Augmentation} Finally, we show that it is important to perform  rebalancing for the training set to achieve good performance. The results without data augmentation ( last row in Table~\ref{comparison_cross_class_correlation}) are significantly worse, due to imbalance.

\begin{table}[!t]
\small
\centering
\caption{Comparison with previous methods in terms of the cross-class $\rho$ on our testing dataset. \textbf{Bold} indicates the best performance.}
\begin{tabular}{P{1cm}||P{1.5cm}|P{1.5cm}|P{2cm}}
\hline
& Noise & Haze & Undesired blur\tabularnewline
\hline\hline
Previous & 0.4199~\cite{liu2013single} & 0.6615~\cite{he2011single} & 0.4864~\cite{chakrabarti2010analyzing} \tabularnewline
Ours & \textbf{0.8174} & \textbf{0.8490} & \textbf{0.6867} \tabularnewline
\hline
\end{tabular}
\label{comparison_previous_methods}
\vspace{-12pt}
\end{table}

\begin{table*}[!t]
\small
\centering
\caption{The performance of our model compared to individual users. The 1st and 2nd rows indicate, for different defects, the average performance of users and our combined multi-column CNN, respectively. \textbf{Bold} indicates best performance over the first two rows. The 3rd row gives a percentage indicating what fraction of  users our model's predictions outperform.}
\begin{tabular}{P{2.3cm}||P{1.2cm}|P{1.4cm}|P{1.3cm}|P{0.9cm}|P{0.9cm}|P{1.3cm}|P{1.55cm}||P{0.9cm}}
\hline
& Bad & Bad white & Over/under & Noise & Haze & Undesired & Bad & Mean\tabularnewline
& exposure & balance & saturation & & & blur & composition & \tabularnewline
\hline\hline
User cross-class $\rho$ & 0.6307 & 0.4953 & 0.6906 & 0.5652 & 0.5755 & \textbf{0.6378} & 0.5348 & 0.5900\tabularnewline
Our cross-class $\rho$ & \textbf{0.7572} & \textbf{0.7217} & \textbf{0.8688} & \textbf{0.6750} & \textbf{0.7391} & 0.6320 & \textbf{0.5990} & \textbf{0.7133}\tabularnewline
\hline
Our ranking & 75\% & 89\% & 100\% & 72\% & 78\% & 44\% & 87\% & 78\%\tabularnewline
\hline
\end{tabular}
\label{user_comparison}
\end{table*}

\vshrink{}
\subsection{Comparison with Previous Methods}
\label{Results of Previous Defect Estimation Methods}

We are not aware of any previous work for simultaneously detecting multiple defects. However, there is previous work for estimating the degree of a single defect, e.g., noise level~\cite{liu2013single} or blur amount~\cite{chakrabarti2010analyzing}. The method in~\cite{liu2013single} can directly predict an overall noise level. The blur estimation method~\cite{chakrabarti2010analyzing} generates a pixel-wise spatially-variant prediction map. We made our best efforts to obtain an overall blur severity assessment from the prediction map, by experimenting with taking different percentiles or the mean. We found that the mean gives the best performance. 

In addition, for some adjustment methods, e.g., haze removal~\cite{he2011single}, we can calculate the adjustment amount for each pixel in the image, where a higher adjustment indicates a more severe \textit{haze} defect in the original image. We can then obtain an overall haze amount estimation by taking the mean adjustment amount over the entire image. Similar to before, we also experimented with various percentiles, but found the mean performed best.

A comparison of our model with these three methods is presented in Table~\ref{comparison_previous_methods}. The cross-rank $\rho$ metric is especially useful here, because the score ranges of these methods are different and not calibrated with our ground-truth scores, but the relative rankings of the test images are  comparable among different methods. We can see that the improvement of our model over these methods is substantial.

\vshrink{}
\subsection{Comparison with Human Performance}

We further compare our predictions to individual users' annotations on the test set. To fairly evaluate a user's performance, we use the mean of the other four users' annotations as ground truth instead of the mean over all the five, to remove the influence of the given user on the ground truth. The same ground truth is then also used to measure the performance of our model, so that the comparison between our model and that user is fair. We show in Table~\ref{user_comparison} the comparison between our results and the averaged user performance. We can see that our model performs better than an average user on most of the defects.

\begin{table}[!t]
\small
\centering
\caption{Our model's performance on five synthesized defects.}
\begin{tabular}{P{1.2cm}|P{1.2cm}|P{1.4cm}|P{1.3cm}|P{0.9cm}}
\hline
Under & Over & Over/under & Gaussian & Motion\tabularnewline
exposure & exposure & saturation & noise & blur\tabularnewline
\hline\hline
0.9560 & 0.8440 & 0.9968 & 0.9573 & 0.8986\tabularnewline
\hline
\end{tabular}
\label{sanity_check}
\end{table}

\vshrink{}
\subsection{Evaluation on Synthetic Data}

Although our model was trained on our dataset of defective images in the wild, we can also validate our trained model on an easier dataset of synthetically generated global defects. We separately generate defective images for under exposure, over exposure, over/under saturation, Gaussian noise, and spatially invariant motion blur. We first select for each defect all of the  defect-free testing images (there are between 420 and 940 such images). For each such image, we synthesize a sequence of defective images with either 11 or 21 different levels of a global parameter, where the number of levels is chosen to be consistent with the class structure in our user dataset discussed in Section~\ref{Data Collection}. We then measure the ranking correlation between the predicted scores and the parameter choices. This can be viewed as a simplification of our cross-class $\rho$, which preserves the three key properties for this task, but does not require random sampling. The mean result over each dataset is listed in Table~\ref{sanity_check}. Note that our model performs better in the synthetic datasets than in the real dataset, which implies that the synthetic task is easier because the defects are global and require less high level information to detect. The result also demonstrates the generalizability of our model. Please see the supplemental material for more details.% of our synthetic adjustments.

% Details of synthetic adjustments (commented out by Connelly to save space)
%We now briefly explain how we generated each synthetic defect. For the exposure defect, we multiplied the intensity by 11 gains. The under-exposure gains have logarithm uniformly spaced in the range $[-1.0, 0.0]$, while over-exposure gains have logarithm uniformly spaced in $[0.0, 1.0]$. For the saturation defect, we scaled the difference between the color and greyscale image by $21$ gains with logarithm uniformly spaced in $[-1.0, 1.0]$. For the noise defect, we added white Gaussian noise: $11$ Gaussian $\sigma$ values are uniformly spaced in $[1/255, 22/255]$. For the motion blur defect, we convolve with $11$ diagonal blur kernels formed by normalizing the first 11 identity matrices. Each synthetic adjustment is applied to between 420 and 940 testing images labelled as defect-free.
%%

\vshrink{}
\subsection{Photographic Defect Localization}

\begin{figure}[!t]
\centering
\includegraphics[width=0.8\linewidth]{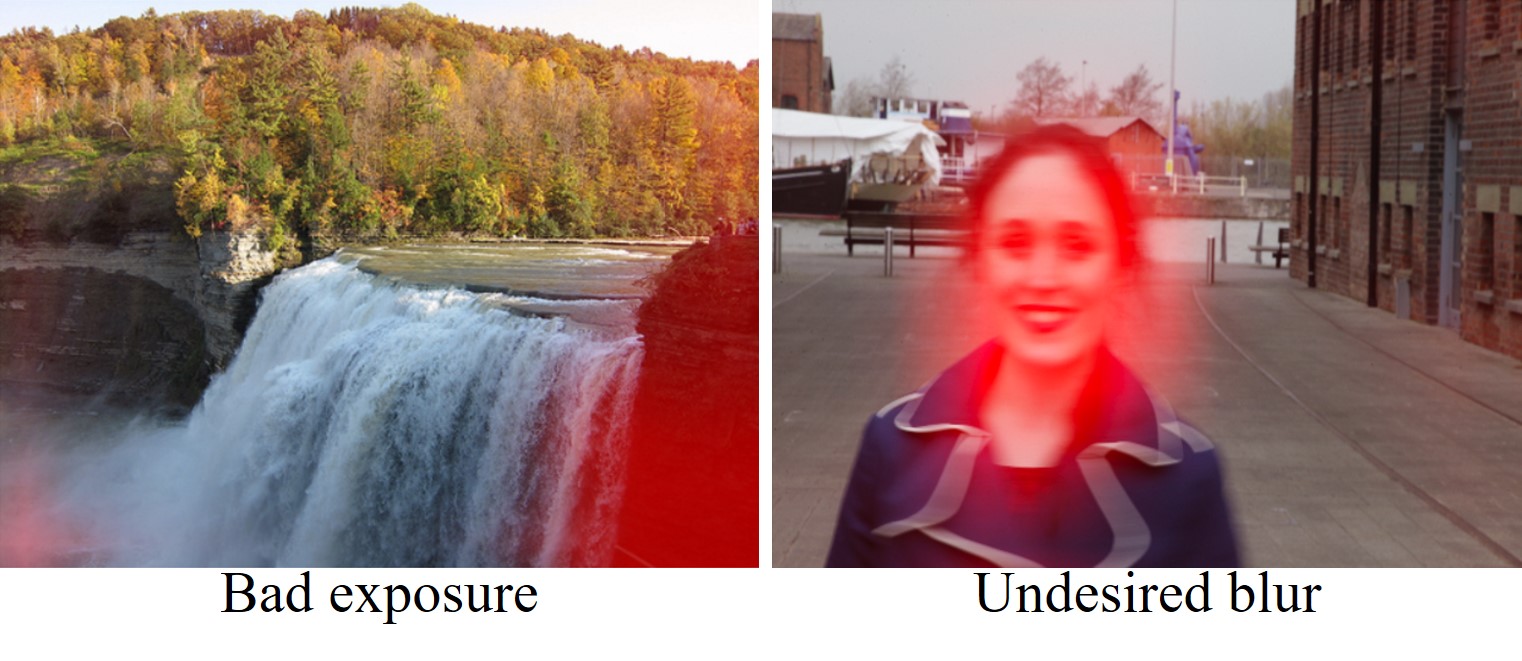}
\caption{Examples of defect localization, where the amount of red color indicates the severity of defects in a local region. In the left image, our heat map highlights indicates that the rock in shadow suffers from the \textit{bad exposure} defect. In the right image, our heat map indicates that the girl's head suffers from motion blur.}
\label{heatmaps}
\vspace{-12pt}
\end{figure}

We also experimented with our well-trained patch model to localize photographic defects. No re-training is required. To do this, we converted the architecture to fully-convolutional~\cite{long2015fully}, by removing the last $7 \times 7$ pooling layer and replacing the fully connected layers with convolutional layers with $1 \times 1$ spatial kernels. We then added an upsampling layer (bilinear interpolation) afterward. The resulting network accepts an image with arbitrary size and outputs a spatially-variant defect map with the same size. Figure~\ref{heatmaps} shows two examples of defect maps. We see our model can roughly localize the defective image areas. Although we have only obtained preliminary results for this, such spatially varying maps could open an avenue for future work such as applications in spatially-variant image corrections or guidance. It could also be a promising future work to collect spatial annotations for defect severity from a user study, and then train a defect localization model specifically.

\vshrink{}
\section{Conclusion}

In this paper, we introduce the problem of simultaneously detecting multiple photographic defects, and make a first attempt of addressing this problem by collecting a large-scale dataset with human annotation, and training a multi-column CNN for prediction. In the experiments, we validated that the proposed model achieves much higher consistency with human judgments than previous single-defect estimation methods as well as baseline CNN models, and also outperforms an average user.% Based on the current work, we plan to further investigate the simultaneous localization and correction of multiple defects in the future.

\section{Acknowledgement}

We thank our anonymous reviewers for beneficial feedback. Thanks to
the photographers for licensing photos under Creative Commons or
public domain. This project was funded by Adobe Research Funding.

\vshrink{}
{\small
\bibliographystyle{ieee}
\bibliography{WACVreferencelist}
}

\end{document}